\documentclass[%
 reprint,
 amsmath,amssymb,
 aps,
]{revtex4-2}

\usepackage{graphicx}
\usepackage{dcolumn}
\usepackage{bm}

\usepackage{amsmath}
\usepackage{url}
\usepackage{color}

\usepackage{here}

\usepackage{algorithmic}
\usepackage{algorithm}

\begin{document}

\title{Artificial Life using a Book and Bookmarker}
\author{Keishu Utimula}
\date{\today}

\begin{abstract}
Reproduction, development, and individual interactions are 
essential topics in artificial life.
The cellular automata, which can handle these in a composite way, 
is highly restricted in its form and behavior 
because it represents life as a pattern of cells. 
In contrast, the virtual creatures proposed by Karl Sims 
have a very high degree of freedom in terms of morphology and behavior. 
However, they have limited expressive capacity in terms of those viewpoints. 
This study carefully extracts the characteristics of 
the cellular automata and Sims models to propose 
a new artificial life model that can simulate reproduction, 
development, and individual interactions while exhibiting 
high expressive power for morphology and behavior. 
The simulation was performed by sequentially reading a book 
with genetic information and repeatedly executing four actions: 
expansion, connection, disconnection, and transition. 
The virtual creatures in the proposed model exhibit 
unique survival strategies and lifestyles and 
acquire interesting properties in reproduction, 
development, and individual interactions 
while having freedom in morphology and behavior. 
\end{abstract}

\maketitle

\section{Introduction}
\label{sec.intro}
In 1994, Karl Sims proposed a system of virtual creatures 
that could be simulated on a computer ~\cite{1994SIM}. 
These virtual creatures had bodies and neurons 
defined by directed graphs, 
and they evolved into creatures 
with various functions through genetic algorithms. 
Subsequently, there have been many related studies on this virtual creature, 
which has a very high degree of freedom 
in its morphology and behavior
 ~\cite{2011LEH, 2012AUE, 2013NIC, 2000LIP, 2003HOR}. 

\vspace{2mm}

However, this virtual creature has limited expressive capacity 
with regard to reproduction, development, and individual interactions. 
Therefore, many prior studies have delved into these elements separately. 

\vspace{2mm}

The process of reproduction is often simulated 
through genetic algorithms. 
Here, there are variations in how the parents are selected 
and how the genes of the selected parents are mixed, 
which are hyperparameters that can be set from the outside ~\cite{2011LEH, 1994SIM2}. 

\vspace{2mm}

The mechanism of biological development 
was applied to a network structure for development  
by Stanley {\it et al.} in 2007 ~\cite{2007STA}. 
In several studies, the bodies of virtual creatures 
were constructed using particles and springs connecting the particles  
to simulate the process of development ~\cite{2009SCH, 2012JOA, 2014JOA, 2015JOA}. 

\vspace{2mm}

Individual interactions have occasionally been 
modeled as parameters for genetic algorithms.
Sims created interesting and diverse virtual creatures 
using the results of competitions between individuals 
as criteria for the selection of the parents ~\cite{1994SIM2}. 
Additionally, some studies have examined changes 
in morphology and behavior in environments 
where predators and prey coevolve ~\cite{2013ITO}.

\vspace{2mm}

These earlier studies independently studied reproduction, 
development, and individual interactions. 
An example of a model that can handle these components in a composite way is the self-reproduction pattern of 
cellular automata proposed by Neumann ~\cite{1966NEU}. 
However, while cellular automata have high expressive power 
for reproduction, development, and individual interactions, 
they are highly restricted in their morphology and behavior 
because they represent life as a pattern of cells. 

\vspace{2mm}

In this study, a new artificial life model is proposed
that can simulate reproduction, development, and individual interactions 
while exhibiting high expressive power for morphology and behavior.
This approach represents an attempt to combine the virtual creatures 
proposed by Sims with the elements possessed by cellular automata.

\vspace{2mm}

The virtual creatures in this proposed model have their 
own survival strategies and lifestyles and have acquired 
interesting properties in reproduction, development, 
and individual interactions.

\vspace{2mm}
\section{Model and Methodology}
\label{sec.det.spec}

A cellular automaton is a discrete computational model 
consisting of a regular grid of cells ~\cite{1966NEU}. 
Each cell changes its internal state according 
to a set of rules. 
In 1966, Neumann discovered the existence of 
a self-reproduction pattern in a cellular automaton 
with 29 internal states ~\cite{1966NEU}. 
Many other researchers followed, including 
Codd ~\cite{2011TIM}, 
Banks ~\cite{1971EDW}, 
and Devore ~\cite{1994JOH}. 
This self-reproduction was not instantaneous, 
but involved a process of development.
In addition, the self-reproduction pattern Loop that 
was proposed by Langton was robust and could achieve 
self-reproduction even when mutations occurred that 
were triggered by collisions with other loops ~\cite{1984CHR, 1999SAY}. 
This is, therefore, an excellent method of expressing reproduction 
that incorporates interactions between individuals. 

\vspace{2mm}

This study focused on the mechanism that causes reproduction and development 
by sequentially reading information about the creatures in genes 
and changing the surrounding environment based on that content. 
This system simulates the expression of various 
reproductive and developmental processes, and even the 
individual interactions caused by the intervention of others.

\vspace{2mm}

The proposed model is based on the concept of incorporating 
cellular automata mechanism into a cell 
that moves freely in a three-dimensional space. 

\vspace{2mm}
\subsection{Overview}
\label{sec.det.spec.01}
The form of the virtual creatures proposed here consists of spheres, 
called cells, and springs that connect them. 
Cells acquire their morphology through repeated cell divisions, 
binding with surrounding cells using springs, or unbinding. 
The procedure is written in the genes that each cell possesses, 
which are read sequentially to represent the developmental process.
The cell can copy some or all of its own genes during division 
and pass them on to a new cell.
In addition, each cell has a neural network with only one hidden layer, 
which allows for a response and action from external input. 
This action is separated from the gene-derived behavior described above. 

\vspace{2mm}

These virtual creatures will live in mutation on an undulating field.
There is a sun above the field that moves back and forth in one direction, 
and light emitting from it is the only source of energy in this world.

\vspace{2mm}
\subsection{Gene}
\label{sec.det.spec.01}
Genes in these virtual creatures are represented as 
strings of 64 different characters. 
Here, a total of 52 upper and lower case letters of the alphabet, 
ten Arabic numerals [$0$-$9$], 
and the symbols + and $-$ were selected as the characters.
Each cell had a string as a gene and a string 
that determined its readout position.
Considering the difference in their roles, 
the gene string is referred to as the Book and the string for the readout position is referred to as the Bookmarker.
For example, the first character to be read is the character "G" if there is a cell with a Book and Bookmarker, as shown 
in Table \ref{table.example}.

\begin{table}[hbtp]
  \caption{Example for the Book and Bookmarker.}
  \label{table.example}
  \centering
  \begin{tabular}{ll}
    \hline
    Book  & ABCD{\color{blue} EF}{\color{red} G}HIJKLMN... \\
    Bookmarker  & {\color{blue} EF} \\
    \hline
  \end{tabular}
\end{table}

The cell can perform four different actions depending 
on the character that is read out first, as follows. 

\begin{itemize}
      \item EXPANSION ($\mathcal{E}$)
      \item CONNECTION ($\mathcal{C}$)
      \item DISCONNECTION ($\mathcal{D}$)
      \item TRANSITION ($\mathcal{T}$)
\end{itemize}

EXPANSION is the generation of new cells around a source cell, 
corresponding to cell division. 
Here, the generated cell is connected to the 
source cell by a spring. 
CONNECTION is the action of connecting 
with surrounding non-connected cells. 
DISCONNECTION is an operation to disconnect 
a cell that has already been connected.
TRANSITION is the action of changing the Bookmarker string.

\vspace{2mm}

In this simulation, 64 different characters were evenly 
assigned to these four actions, as listed in Table \ref{table2}.
\begin{table}[hbtp]
  \caption{
  Characters assigned to the four actions.}
  \label{table2}
  \centering
  \begin{tabular}{ll}
    \hline
    $\mathcal{E}$ &:  ABCDEFGHIJKLMNOP\\
    $\mathcal{C}$ &:  QRSTUVWXYZabcdef\\
    $\mathcal{D}$ &:  ghijklmnopqrstuv\\
    $\mathcal{T}$ &:  wxyz0123456789+$-$\\
    \hline
  \end{tabular}
\end{table}

Thus, EXPANSION is the action that would have been performed 
in the previous example, as shown in Table \ref{table3}.

\begin{table}[hbtp]
  \caption{
  Examples for EXPANSION.}
  \label{table3}
  \centering
  \begin{tabular}{ll}
    \hline
    Book  & ABCD{\color{blue} EF}{\color{red} G}HIJKLMN... \\
    Bookmarker  & {\color{blue} EF} \\
    Action & [{\color{red} G}] EXPANSION\\
    \hline
  \end{tabular}
\end{table}

In terms of interpretation of actions, there is no difference 
between characters classified as the same action.
Thus, in the example here, the behavior 
would be the same even if G were A or P.
The action determines how the string that follows is interpreted.

\subsubsection{EXPANSION}
\label{sec.det.spec.01}

The following string is interpreted as information about the newly created cell if the action is EXPANSION.
Specifically, the information is as follows. 

\begin{itemize}
  \item Light absorption rate (color)
  \item Luminous intensity
  \item Mass
  \item Radius
  \item Neural network weight
  \item From where to where in the Book to be copied and passed to the new cell
  \item Bookmarker for new cells
\end{itemize}

The cell generates new cells based on this information.
The location to be generated is around the original cell, but it is random.
Additionally, the generated cell is connected with 
the cell from which it was generated.
Any further strings following this are interpreted 
as the next Bookmarker in this cell.
Specifically, the following string was read by every other 
character and used as the new Bookmarker.
For example, the new Bookmarker would be "df" 
if the length of the Bookmarker was 2 and 
the string following the specifics of the action was 
"def...", 
as shown in Table \ref{table.expansion}.
This is a simple encryption to prevent the instruction 
about the new Bookmarker itself from becoming the next read position. 

\begin{table}[H]
  \caption{
  Examples for EXPANSION.}
  \label{table.expansion}
  \centering
  \begin{tabular}{ll}
    \hline
    Book  & CDEF{\color{red} A}\textbf{HIJK...abc}{\color{blue} defg...} \\
    Bookmarker  & EF \\
    Action & [{\color{red} A}] EXPANSION\\
    New Cell Info. & \textbf{HIJKLMN...abc}\\
    New Bookmarker  & {\color{blue} df}\\
    \hline
  \end{tabular}
\end{table}

\vspace{2mm}
\subsubsection{CONNECTION}
\label{sec.det.spec.01}

The cell will be 
in a waiting state for the connection if the action is CONNECTION. 
This is the state in which a connection will be performed 
if there is a cell near another cell that is also waiting to be connected. 
The following string is interpreted as the next Bookmarker, as shown in
Table \ref{table.connection}. 

\begin{table}[H]
  \caption{
  Examples for CONNECTION.}
  \label{table.connection}
  \centering
  \begin{tabular}{ll}
    \hline
    Book  & ABCDEF{\color{red} Q}{\color{blue} defg...} \\
    Bookmarker  & EF \\
    Action & [{\color{red} Q}] CONNECTION\\
    State & Waiting for connection\\
    New Bookmarker  & {\color{blue} df}\\
    \hline
  \end{tabular}
\end{table}

\vspace{2mm}
\subsubsection{DISCONNECTION}
\label{sec.det.spec.01}

The cell will be in a state waiting for disconnection if the action is DISCONNECTION.
This is the state in which a cell will be disconnected 
if a connected cell is also waiting to be disconnected. 
The following string is interpreted as the next Bookmarker, as shown in
Table \ref{table.disconnection}. 

\begin{table}[H]
  \caption{
  Examples for DISCONNECTION.}
  \label{table.disconnection}
  \centering
  \begin{tabular}{ll}
    \hline
    Book  & ABCDEF{\color{red} g}{\color{blue} defg...} \\
    Bookmarker  & EF \\
    Action & [{\color{red} g}] DISCONNECTION\\
    State & Waiting for disconnection\\
    New Bookmarker  & {\color{blue} df}\\
    \hline
  \end{tabular}
\end{table}

\vspace{2mm}
\subsubsection{TRANSITION}
\label{sec.det.spec.01}

The same process as described so far is followed if the action is TRANSITION,  
where the following string is interpreted as the next Bookmarker. This is shown in
Table \ref{table.transition}. 

\begin{table}[H]
  \caption{
  Examples for TRANSITION.}
  \label{table.transition}
  \centering
  \begin{tabular}{ll}
    \hline
    Book  & ABCDEF{\color{red} w}{\color{blue} defg...} \\
    Bookmarker  & EF \\
    Action & [{\color{red} w}] TRANSITION\\
    New Bookmarker  & {\color{blue} df}\\
    \hline
  \end{tabular}
\end{table}

\vspace{2mm}
\subsubsection{Shift for Bookmarker}
\label{sec.det.spec.01}

It is easy to realize 
a procedure that repeats the same action indefinitely using the implementation described above. 
An example of the Book and Bookmarker that repeatedly expand 
the same cell is shown in Table \ref{table.repeat}.

\begin{table}[H]
  \caption{
  Example of Book and Bookmarker 
  with the repeated EXPANSION of the same cell.}
  \label{table.repeat}
  \centering
  \begin{tabular}{ll}
    \hline
    Book  & CDEF{\color{red} A}\textbf{HIJK...abc}{\color{blue} EeFghi...} \\
    Bookmarker  & EF \\
    Action & [{\color{red} A}] EXPANSION\\
    New Cell Info. & \textbf{HIJKLMN...abc}\\
    New Bookmarker  & {\color{blue} EF}\\
    \hline
  \end{tabular}
\end{table}

However, if the same operation is to be repeated 
a finite number of times, the same content must be described 
in the Book for that number of times, which is inefficient.
Therefore, a string called Advance was introduced 
to eliminate this problem. 
This Advance string shifts the readout position of the new Bookmarker.
For example, the new Bookmarker would be "Fh" if the Advance was 2 in Table \ref{table.repeat}.
The Advance was also updated at this time. Subsequently, 
the string that followed the string interpreted 
as the new Bookmarker became the new Advance.
In this example, the new Advance is "i" if the length of the Advance is one.
This Advance string is interpreted 
as a 64-decimal integer at runtime. 

\vspace{2mm}

\begin{figure}[htbp]
    \centering
    \includegraphics[width=1.0\hsize]{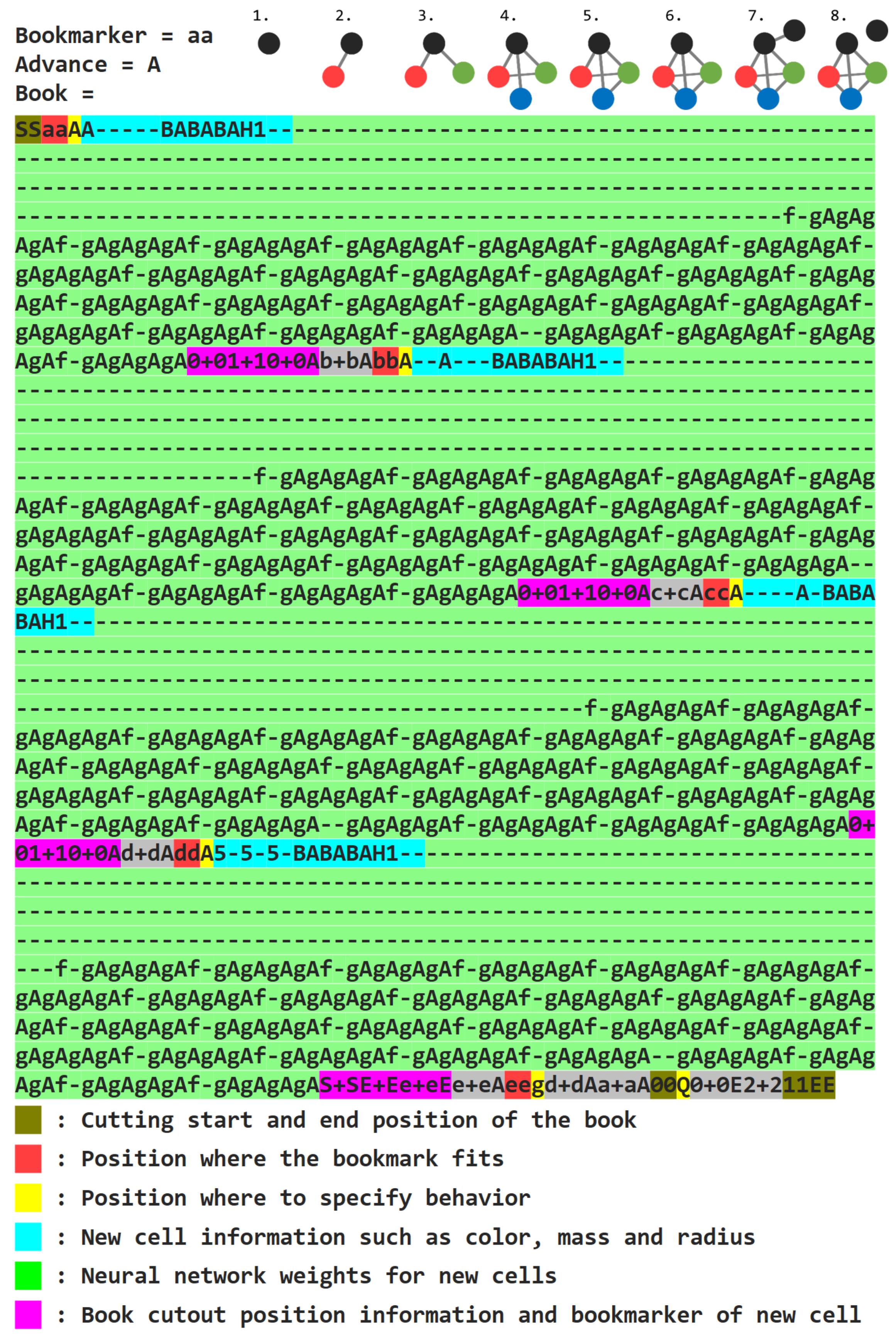}
    \caption{
    Book repeatedly self-reproducing while forming tetrahedral bodies.
    }
    \label{fig.14}
\end{figure}

As an example, the Book that repeats 
self-reproduction while forming a tetrahedral body 
can be written based on the above, as shown in Fig. \ref{fig.14}. 

\vspace{2mm}
\subsection{Neural Networks}
\label{subsec.nn}

Each cell has a neural network with only one hidden layer.
The weights of this neural network were determined 
by the genes and were not variable.
Additionally, the activation function was $\tanh$.

\vspace{2mm}

The output of this neural network is as follows:

\begin{itemize}
  \item Output $S_{out}$ to connected cells 
  \item Output $L_{out}$ to connected cells 
  \item Output $E_{out}$ to connected cells 
  \item Determine whether or not to read the Book 
  \item Determine the proportion of the contacted cell to be eaten
  \item Determine whether or not to incorporate the Book of contacted cells
  \item Determine whether or not to emit light
\end{itemize}

Here, $L_{out}$ is the output for changing the position 
relative to the connected cell.
The natural length of the springs connecting each cell increases 
if the value is positive and magnitude is greater than 
a certain threshold value. 
Conversely, the natural length of the spring decreases 
if the value is negative and magnitude is greater than 
a certain threshold. 
Therefore, this mechanism functions as a muscle 
in this virtual creature. 
Additionally, this value is used as input to the neural network 
of the connected cell. 

\vspace{2mm}

$E_{out}$ determines the amount of energy sent to the connected cell.
The cell transports its own energy to the 
connected cell at a rate corresponding to the magnitude of 
the value if this value is positive. 
This value is also used as an input to the 
neural network of the connected cell.

\vspace{2mm}

In contrast, $S_{out}$ causes no action and its value is 
simply used as input to the connected cell.
This allows each cell to communicate with connected cells. 
The input to this connected cell is multiplied 
by the variable $S'$ and not simply the value of 
the output $S$ as is. 
The variable $S'$ is updated according to 
the value of the output $S$, which is expressed as: 
\begin{align}
S'\left(t+\Delta t\right) 
= S'\left(t\right) + {\Delta s} \times S\left(t\right)
\label{eq.hebb}
\end{align}
This system functions as variable weights in this neural network.
In other words, $S'$ is the coupling strength of the neural network 
between the connected cells, 
on which the feedback from Eq. (\ref{eq.hebb}) is applied. 
This is a simplified version of Hebb's rule, 
in which the firing of a neuron improves the strength of 
the connections between neurons ~\cite{1949HEB}.
Here, ${\Delta s}$ is a parameter that determines 
the simulation environment. 
This parameter was set to $0.1$ for the simulations in this study.

\vspace{2mm}

The inputs to this neural network are as follows: 

\begin{itemize}
  \item Output $S_{out}$ from the connected cell
  \item Output $L_{out}$ from the connected cell
  \item Output $E_{out}$ from the connected cell
  \item Ratio of red in light hitting the cell
  \item Ratio of green in light hitting the cell
  \item Ratio of blue in light hitting the cell
  \item Information that other cells have been contacted
\end{itemize}

\begin{figure}[htbp]
    \centering
    \includegraphics[width=1.0\hsize]{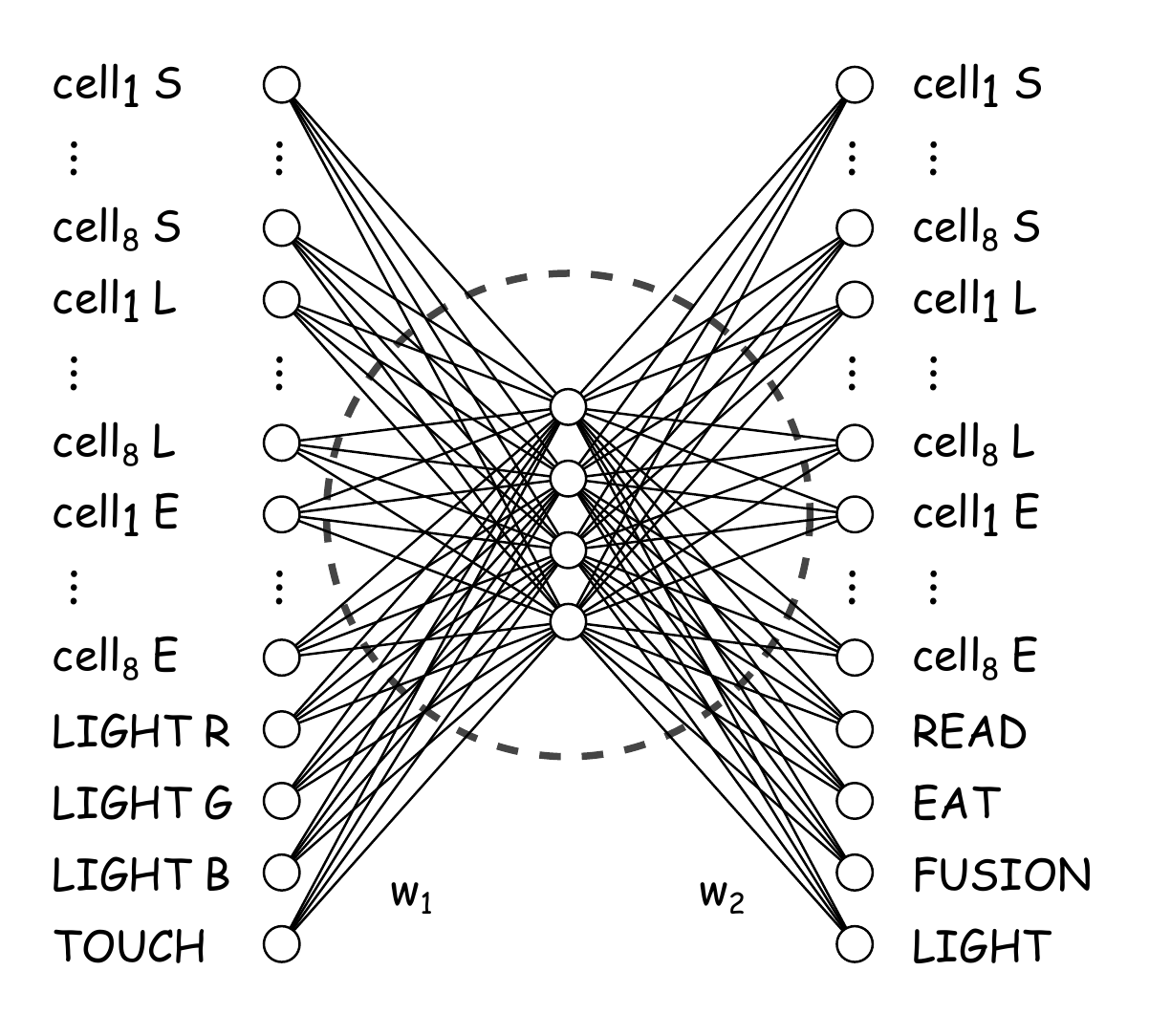}
    \caption{
    Schematic diagram of a neural network in one cell.
    The left and right sides are the input and output, respectively.
    ${\rm cell}_X$ denotes the $X$th connected cell, 
    and $E$ and $L$ denote the energy and natural length of the spring, 
    respectively. 
    LIGHT and TOUCH on the input side are 
    the light and cell information in contact with the cell, 
    respectively.
    On the output side, READ determines whether to read the Book, 
    EAT determines whether to eat the contacted cells, 
    FUSION determines whether to take in the Book of the contacted cells, 
    and LIGHT determines whether to emit light.
    }
    \label{fig.nn}
\end{figure}

A schematic diagram of this neural network is 
shown in Fig. \ref{fig.nn}.

\vspace{2mm}
\subsection{Energy}
\label{sec.det.spec.01}
Cells have a quantity called energy, whose value decreases 
at a fixed rate at each simulation step. 
The cell will die if the value falls below a certain amount, 
which mimics the metabolism of an organism. 

\vspace{2mm}

The only source of energy in this world is light from 
the sun placed above the field, and virtual creatures 
cannot increase their energy by themselves. 
A cell can absorb light at a rate corresponding 
to its color and retain it as energy. 

\vspace{2mm}

There are three situations in which the energy of a cell increases. 

\begin{itemize}
  \item When absorbing light
  \item When eating other cells
  \item When receiving energy from a connected cell
\end{itemize}

Conversely, there are five situations 
where cell energy is reduced. 

\begin{itemize}
  \item When time has passed
  \item When eaten by other cells
  \item When sending energy to the connected cell
  \item When generating a new cell
  \item When emitting light
\end{itemize}

\vspace{2mm}
\subsubsection{Energy consumption over time}
\label{subsec.decay}

As mentioned earlier, the cell's energy decreases 
with each simulation step. 
The cell energy $E$ after $\Delta t$ 
(one step) is updated, which is expressed as: 

\begin{align}
E\left(t+\Delta t\right) 
&= 
E\left(t\right) - 
\left(
C \cfrac{A^{N_c}}{A^{N_{\rm max}} }
\right)
E\left(t\right) {\Delta t}\nonumber\\
&=
E\left(t\right) - 
U
E\left(t\right)
{\Delta t}
\end{align}

where $N_{\rm max}$ is the maximum number of cells 
that can be connected to the cell and $N_c$ is the number 
of cells that are currently connected to the cell.
In addition, $C$ and $A$ are parameters for adjusting 
energy consumption, where $C$ is a real number greater than 
$0$ and $A$ is a real number greater than $1$.
In this simulation , $1.0$ was used for $C$, 
and $A$ was adjusted in real-time so that the number of cells 
in the field was $4000$.
This formula shows that more energy is consumed when more cells are connected.

\vspace{2mm}

In this simulation, there is no energy expenditure 
due to exercise.
Here, exercise means changing the natural length 
between connected cells.
Also, energy consumption is not considered for exchanging 
information between cells using neural networks.
In practice, however, it is more realistic to consider 
this type of energy consumption.
The energy consumption, according to the number of 
connected cells mentioned above, is introduced to easily
take this energy consumption process into account. 

\vspace{2mm}
\subsubsection{Energy change when eating cells }
\label{subsec.eat}

Cells can eat the cells they are in contact with, 
but two rules apply. 
The first rule is that if the number of connections of the cell 
to be eaten is equal to or greater than its number of connections, 
the cell cannot be eaten. 
The second rule is that if the number of connections of 
the cell that is about to be eaten is smaller than its own, 
the more energy it can obtain at one time. 

\vspace{2mm}

The equation for the energy $\Delta E$ 
obtained by eating a cell is expressed as:

\begin{align}
\Delta E = \left\{
\begin{array}{ll}
d\cfrac{N_c-N'_c}{{N_{\rm max}}} & (N_c > N'_c)\\
0 & (N_c \leq N'_c)
\end{array}
\right.
\end{align}

where $d$ is the output of the aforementioned neural network. 
Moreover, it is simply $0$ when it is less than $0$. 
In addition, $N_c$ is the number of connections of cells attempting 
predation and $N'_c$ is the number of connections of cells about 
to be predated.

\vspace{2mm}
\subsubsection{Energy change when a new cell is generated}
\label{sec.energy4generating}

Energy is required for one cell to generate a new cell.
There are two perspectives on this "required energy".
One perspective is that the energy is consumed by the generation of the cell, 
and the other is that the energy is passed on to the newly generated cell.
It is reasonable to assume that the required energy is the sum of these two perspectives.
However, the energy consumed by the generation will not be considered in this simulation.
Thus, the only energy that the cell loses in generating a new cell 
is the energy it passes to the new cell at the time of generation.
Here, the magnitude of energy passed to a new cell is based on 
the energy that the generated cell could acquire 
after a sufficient amount of time.

\vspace{2mm}

When a cell obtains energy from only one light source, 
the obtained energy converges to $E_{\infty}$ 
in a certain amount of time; 
$E_{\infty}$ can be estimated as follows:

\begin{align}
E_{\infty} = 
\cfrac{\Delta E_L}{1-\exp \left[ -\cfrac{U}{P\Delta S} \right]}
\label{eq.e_inf}
\end{align}

where $\Delta E_L$ is the energy obtained from one photon, 
$P$ is the amount of light received per unit time and unit area, 
and $\Delta S$ is the area over which the cell receives light.
It is also assumed that the distance between a light source 
and cell does not change.
Details of this derivation are provided in the Appendix.

\vspace{2mm}

Using Eq. (\ref{eq.e_inf}), the energy 
$E_{\infty}^{\rm sun}$ that can be reached could be predicted if the new cell to be 
generated was only supplied with energy from the sun.
This energy that was multiplied by a factor was defined as the amount of 
energy that should be passed to the cell during generation. 
This implies that cells that obtain more energy 
in the future will have higher generation costs. 
This factor was set to 0.5 in the simulations performed here.

\vspace{2mm}
\subsection{Environment}
\label{sec.det.spec.01}

\vspace{2mm}
\subsubsection{Field}
\label{sec.det.spec.01}

\begin{figure}[htbp]
    \centering
    \includegraphics[width=1.0\hsize]{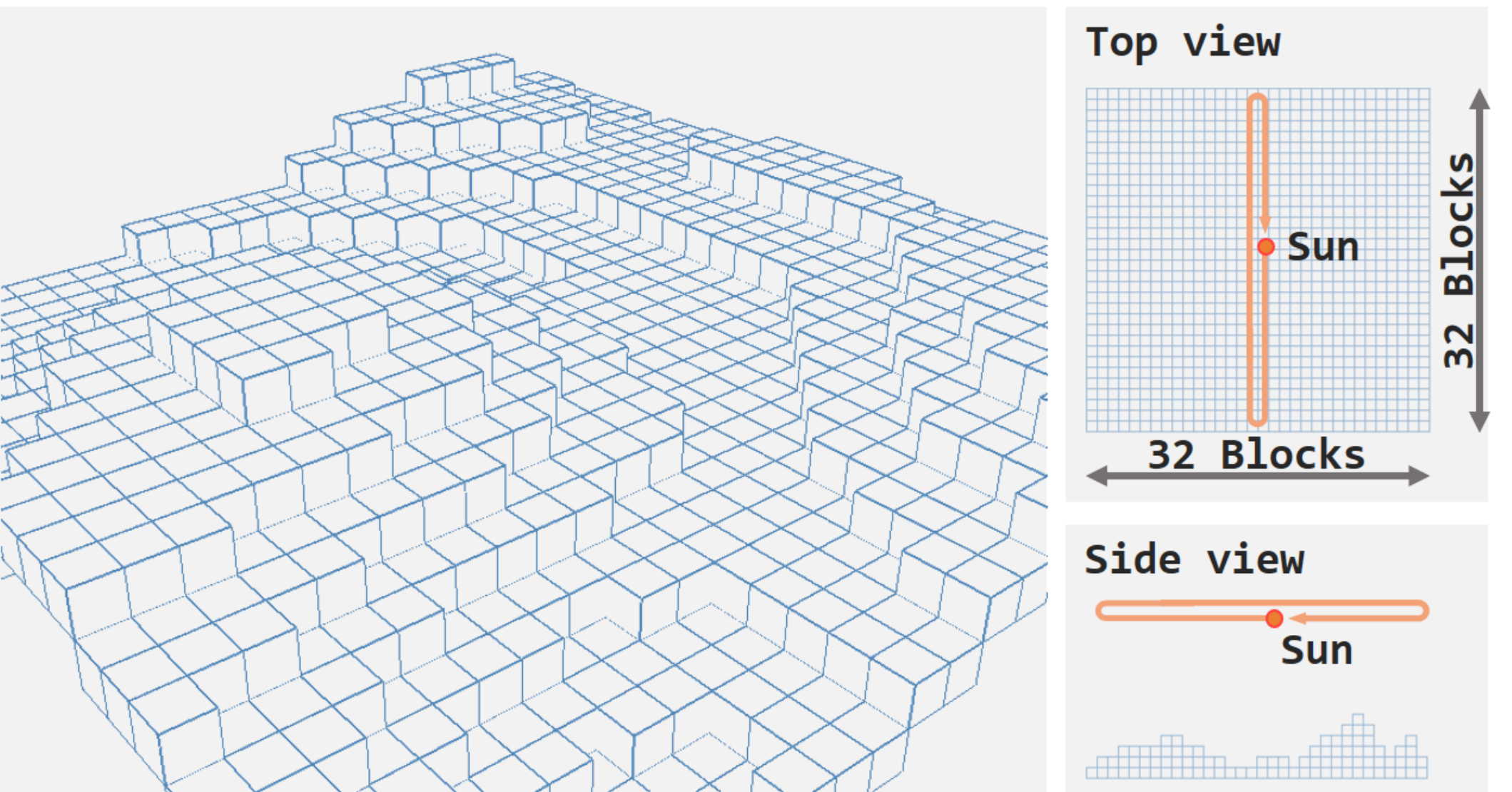}
    \caption{
    Example of a field (left), top view (right-top), 
    and side view (right-bottom). 
    The sun is placed at the top of the field, 
    moving back and forth in one direction.
    }
    \label{fig.19}
\end{figure}

Simulations were performed on a field of 
$32\times 32$ blocks containing undulations. 
The maximum radius of the cell was set to approximately $0.16$ 
when the length of one side of this block was $1$.
One example of a field is shown in Fig. \ref{fig.19}.
The sun, which moves back and forth in one direction, 
is located at the top of the field, and the light emitted 
from it is the only source of energy in this field.

\vspace{2mm}
\subsubsection{Mutation}
\label{sec.det.spec.01}

There are two types of mutations:
one mutation can occur for all cells at every simulation step, 
and the second mutation occurs only for the Book of new cells 
during cell division. 
These mutation probabilities are represented by 
$\alpha$ and $\beta$, respectively.
The mutation occurring for all cells at every simulation step 
may mutate against the Book or Bookmaker.

\vspace{2mm}

\begin{figure}[htbp]
    \centering
    \includegraphics[width=1.0\hsize]{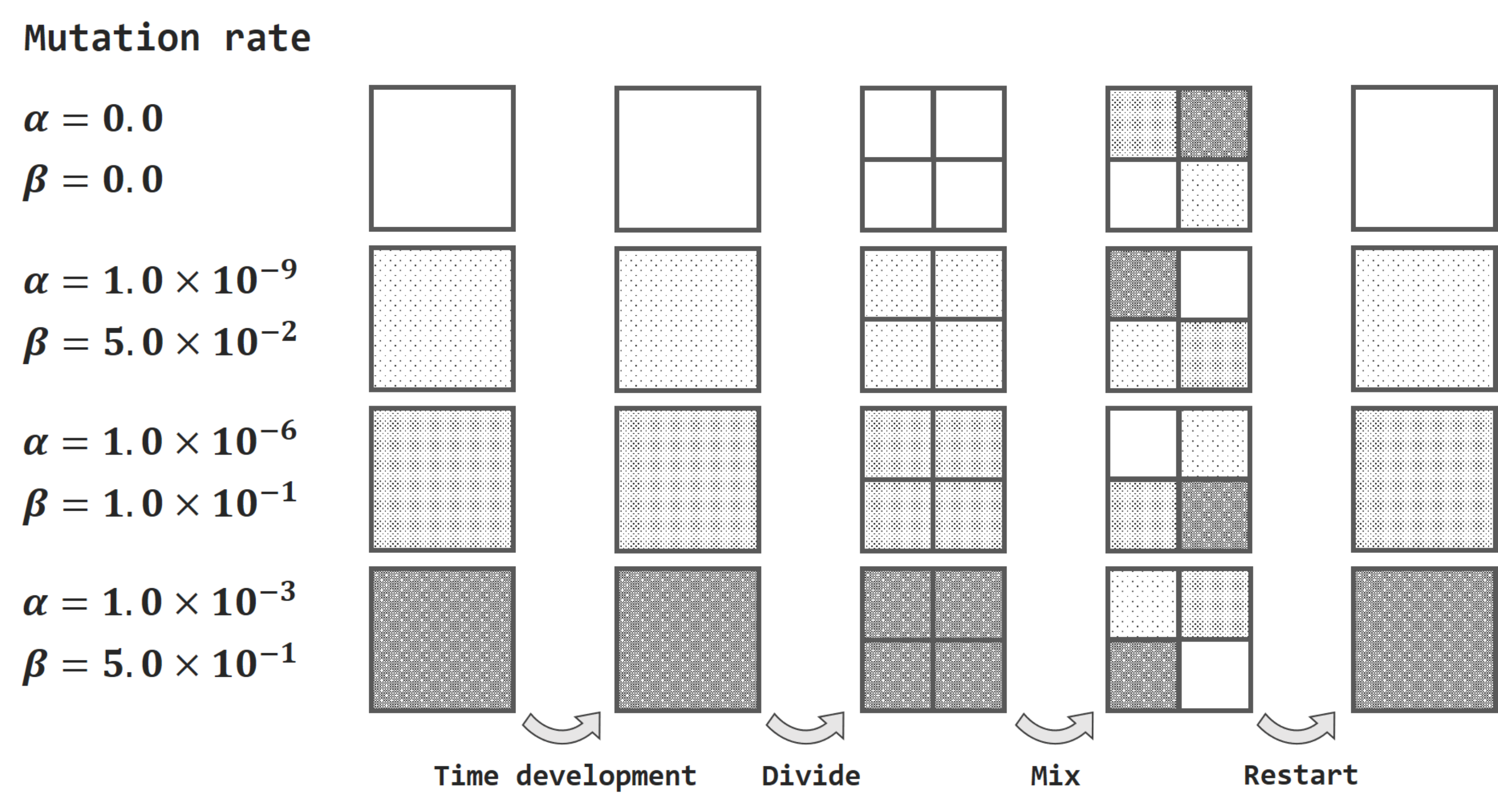}
    \caption{
    Conceptual diagram of mutation.
    The probability of a mutation occurring for all cells at each step of 
    the simulation is $\alpha$, and the probability of a mutation occurring 
    only for the new cell Book at cell division is $\beta$. 
    Four fields were prepared and assigned different mutation 
    probabilities to each field. 
    After simulating a specific amount of time in each field, 
    the virtual creatures grown in each field are distributed to the 
    four fields and the simulation is rerun.
    }
    \label{fig.15}
\end{figure}

Four fields were independently calculated 
with different mutation probabilities in this simulation and 
the virtual creatures were mixed on each field after a specific time, 
as shown in 
the conceptual diagram in Fig. \ref{fig.15}.
The length of the specific time (Time development) was 
set to $10^6$ steps.

\vspace{2mm}
\subsubsection{Efficiency for light energy conversion}
\label{sec.color}

The amount of light absorbed by a cell is expressed as the product of 
the light intensity and light absorption coefficient of the cell, 
which consists of three components: red, green, and blue.
In other words, 
if the light intensity is $I_r$, $I_g$, and $I_b$ for each component 
and the respective absorptivities of the cell are $a_r$, $a_g$, and $a_b$, 
then the amount of light $\Delta L$ absorbed by the cell can be expressed as: 
\begin{align}
\Delta L = I_r a_r + I_g a_g + I_b a_b \quad .
\end{align}
Here, the parameters of how much of each color 
is converted to energy must be determined. 
The conversion efficiencies of the three components into energy are expressed 
as $c_r, c_g,$ and $c_b$. Subsequently, the amount of energy $\Delta E_L$ 
obtained from the light absorbed by the cell is calculated using:
\begin{align}
\Delta E_L = c_r I_r a_r + c_g I_g a_g + c_b I_b a_b \quad .
\end{align}
The parameters in this simulation were set to 
$c_r = 0.8, c_g = 0.2,$ and $c_b = 0.8$.
This approach is also used when determining the cost of generating cells, 
as described in Section \ref{sec.energy4generating}.
For example, 
white and black cells have smaller and higher generation costs, 
respectively, as indicated from the above equation.
Therefore, this parameter is expected to affect the color of future cells.

\vspace{2mm}
\subsubsection{Parameters for connections}
\label{sec.conn}

Cells can connect to surrounding cells, but it is unnatural 
to connect to cells that are too far apart. 
Therefore, only cells within $1.95$ times 
their radius $r$ were allowed to be connected. 
In addition, the natural length of the connections between cells 
were varied according to the output of the neural network, 
but this is also unnatural if it is stretched too long. 
Therefore, this limit was set to $1.10$ times a cell’s own radius $r$.
Lastly, the connections will be broken if the distance between 
connected cells is stretched for some reason and 
exceeds $2.00$ times the natural length.

\vspace{2mm}
\section{Results and Discussion}
\label{sec.results}

A virtual creature was simulated for 
self-reproduction that formed a tetrahedron as its first 
virtual creature, as shown in Fig. \ref{fig.14}. 
Additionally, the kind of virtual creatures that emerged was observed. 

\vspace{2mm}

The results of the investigation of the ecology of 
the two observed virtual creatures are presented here, 
as well as a discussion of the results.

\vspace{2mm}
\subsection{Dumbbell-shaped virtual creatures}
\label{subsec.dumbbell}

\begin{figure}[htbp]
    \centering
    \includegraphics[width=1.0\hsize]{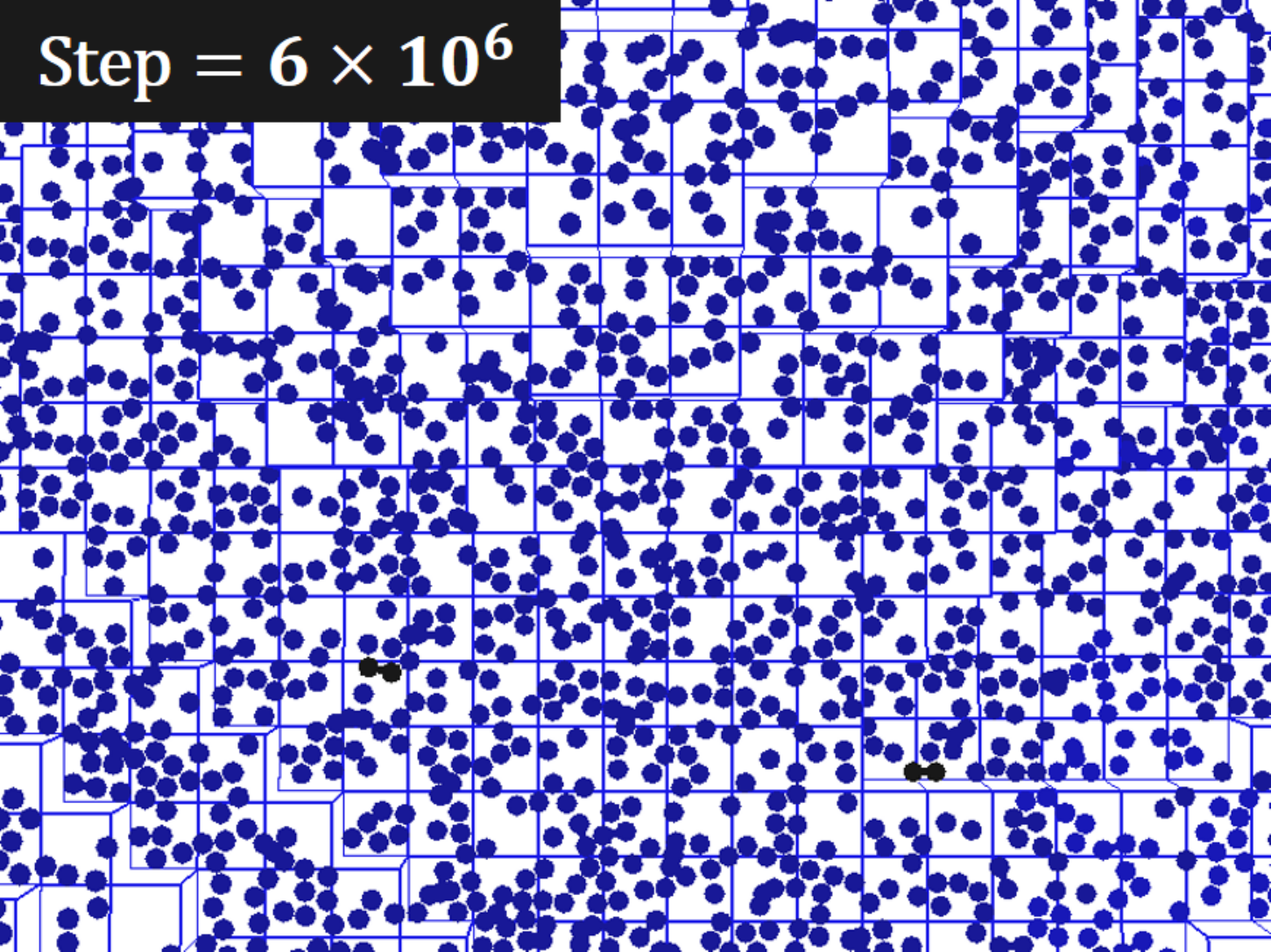}
    \caption{
    Example of a single cell proliferating 
    by simple division.
    }
    \label{fig.02}
\end{figure}

In the simulations performed, 
single cells were generated in many cases that were proliferated 
by simple division in the early stages. 
One example of this is shown in Fig. \ref{fig.02}.

\vspace{2mm}

\begin{figure}[htbp]
    \centering
    \includegraphics[width=1.0\hsize]{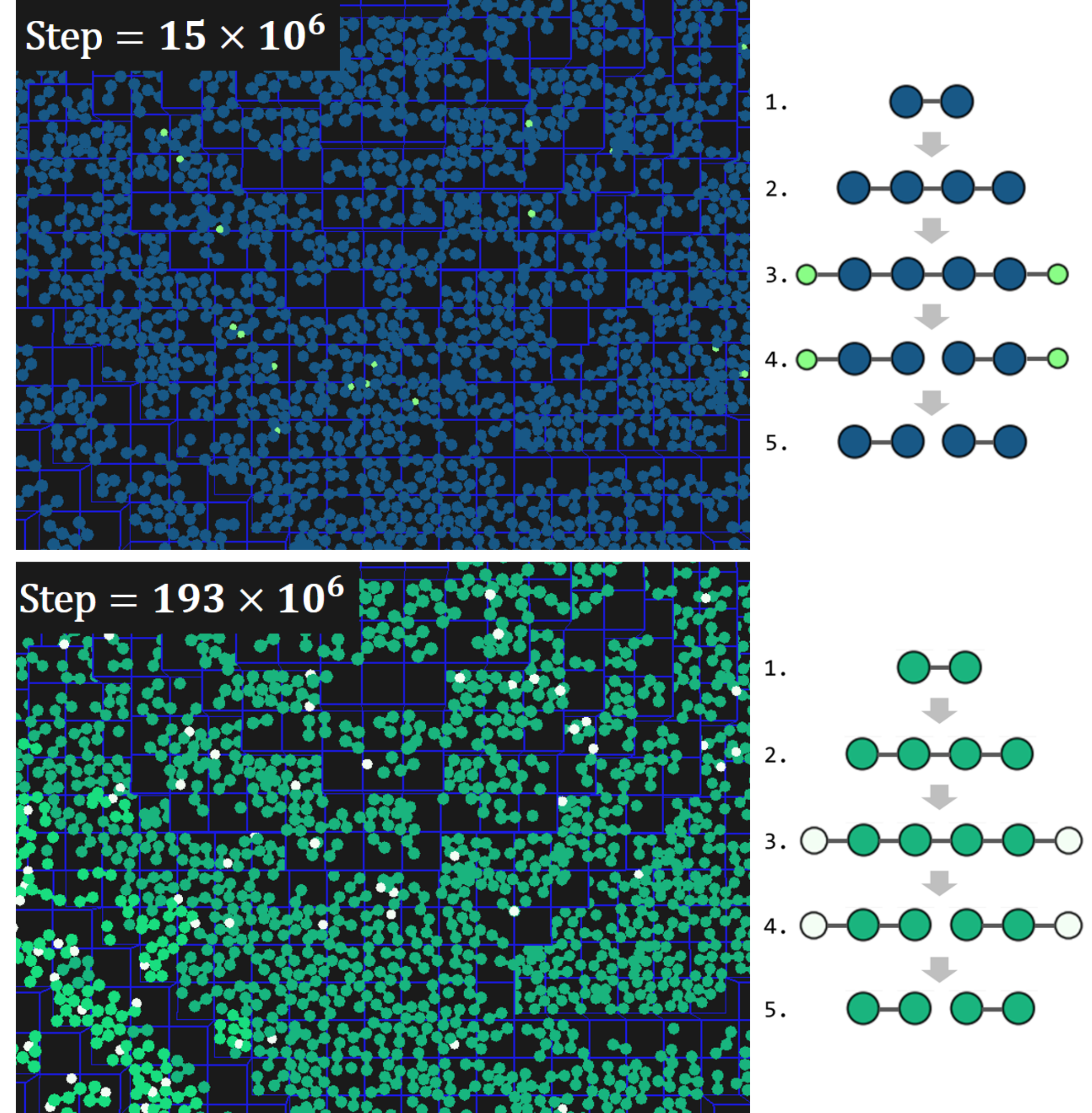}
    \caption{
    Dumbbell-shaped virtual creatures evolved from a single cell. 
    The actual appearance of the virtual creature is shown on the left, 
    and a schematic of the self-reproduction is shown on the right.
    }
    \label{fig.04}
\end{figure}

Dumbbell-shaped virtual creatures could be seen after continuing the simulation for some time from this point, as shown in Fig. \ref{fig.04}.
The actual appearance of the virtual creatures is shown on the left, 
and a schematic of the self-reproduction is shown on the right. 
This dumbbell-shaped morphology was first seen at approximately 
$15\times 10^{6}$ steps after the start of the simulation 
(top row in Fig. \ref{fig.04}). 
At the $193\times 10^{6}$ step, which was more than $10$ times longer, 
the morphology had not changed significantly, merely 
in color and size (bottom row in Fig. \ref{fig.04}). 
As mentioned in Section \ref{sec.color}, the energy conversion efficiency was 
set high for red and blue light, so this color change was considered to be 
an adaptation that reduced the absorption rate of green, 
which had a low energy conversion efficiency.

\vspace{2mm}

Focusing on the morphogenesis process of this dumbbell-shaped virtual creature, 
the white cells were born between the second and third steps 
and disappeared between the fourth and fifth steps.
This disappearance was caused by the green cells that were connected to 
white cells eating the white cells. 
It was considered that this seemingly inexplicable process, 
in which the green cells ate the cells they generated, 
prevented the green cells from being eaten by other virtual creatures. 
As discussed in Section \ref{subsec.eat}, cells with a high number of connections 
can eat cells with a low number of connections.
Thus, without this white cell, the green cell at the edge would be 
vulnerable to predation by other virtual creatures 
because it only has one connection.
This white cell increases the number of connections of the green cell 
and works as a protector to physically prevent other virtual creatures 
from approaching. 
In fact, 
the protector is changing to white and large, 
as shown in Fig. \ref{fig.04}. 
This is a reasonable change when trying to expand the protector 
to keep predators away, at the lowest possible generation cost.

\vspace{2mm}

\begin{figure}[htbp]
    \centering
    \includegraphics[width=1.0\hsize]{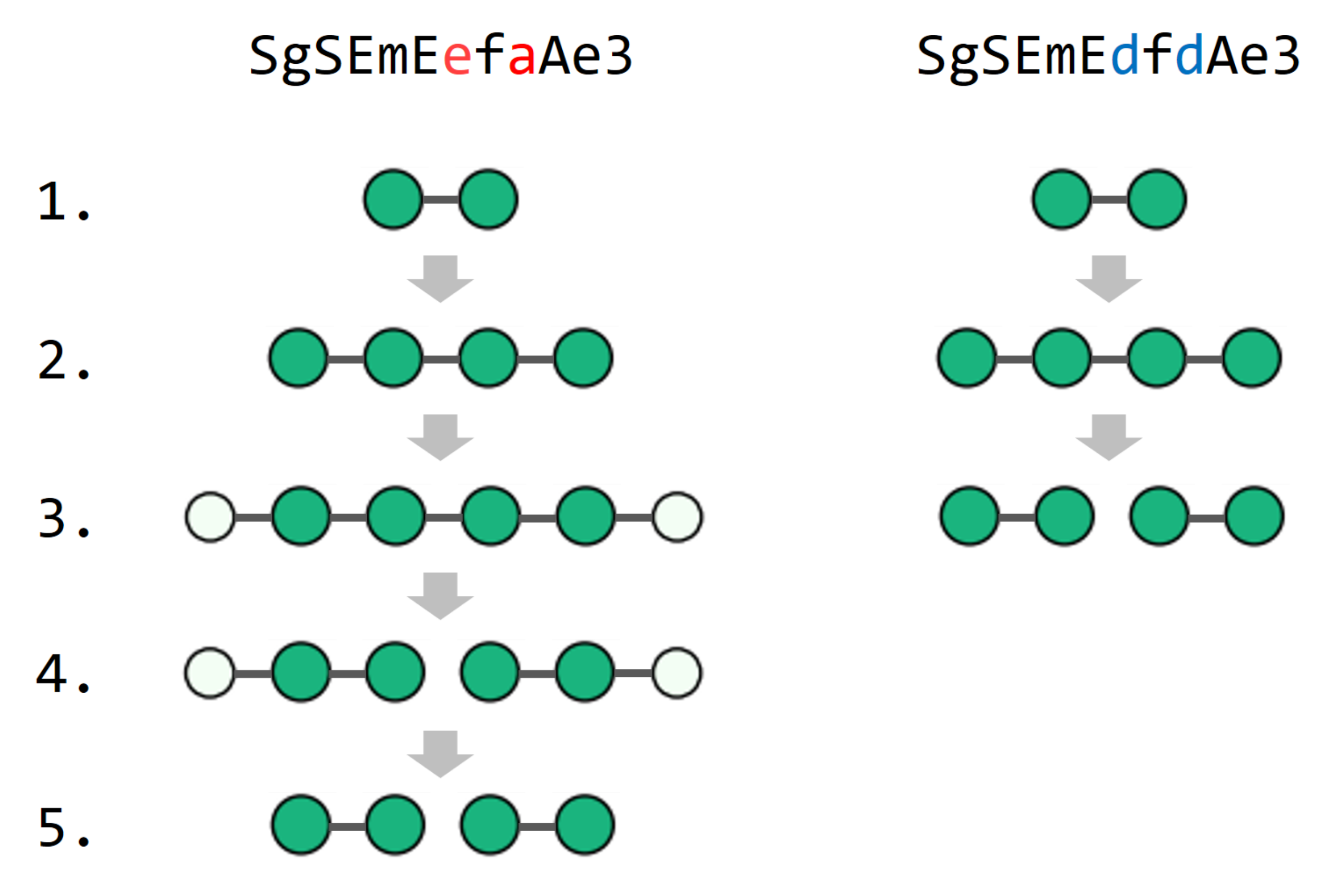}
    \caption{
    Comparison of the morphogenesis process of a dumbbell-type 
    virtual creature (left) and a virtual creature whose Book was 
    edited to remove the process of protector generation (right). 
        }
    \label{fig.05}
\end{figure}

To confirm this role, a virtual creature was prepared
by partially editing the Book of this dumbbell-shaped virtual creature 
to remove the process that generated the white cells.
Specifically, "SgSEmEefaAe3" in the Book was edited 
to "SgSEmEdfdAe3". 
The differences in this morphogenetic process 
are shown in Fig. \ref{fig.05}.
This virtual creature with omitted protector generation had a superior proliferation rate compared to 
the original dumbbell-type virtual creature 
because of its short self-reproduction cycle. 

\vspace{2mm}

\begin{figure}[htbp]
    \centering
    \includegraphics[width=1.0\hsize]{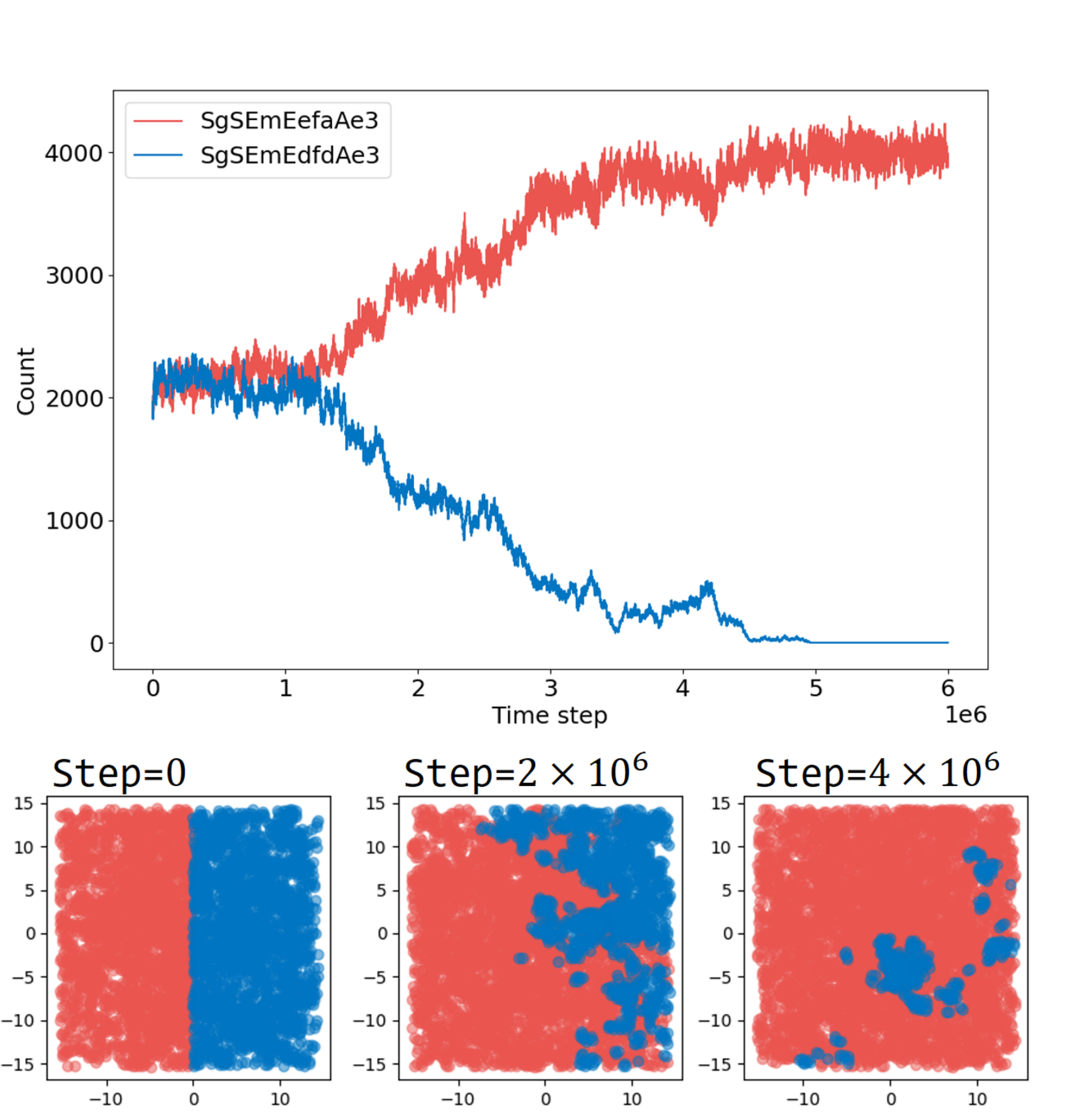}
    \caption{
    Population changes of a dumbbell-shaped virtual creature with 
    a protector (red) and virtual creature omitting the process of 
    protector development (blue). 
    The top panel shows the change in the respective populations 
    with time evolution, while the bottom panel visualizes the distribution 
    in the field for the three time periods. 
    }
    \label{fig.06}
\end{figure}

These two types of virtual creatures were placed in the same field 
and changes to their respective populations were observed, as shown in Fig. \ref{fig.06}.
Note that the simulation was run with a fixed $A$, 
referring to the simulation results of the dumbbell-shaped 
virtual organism before editing. 
The role of parameter $A$ was discussed in Section \ref{subsec.decay}. 
As time passes, the populations for virtual creatures 
without protectors will decline and eventually become extinct. 

This result strongly suggests that the white cells function as protectors. 
It was considered that the morphology was acquired from the single-celled state 
shown in Fig. \ref{fig.02}, reflecting individual interactions. 

\vspace{2mm}
\subsection{Reticulated virtual creatures}
\label{sec.det.spec.01}

The dumbbell-shaped virtual creatures discussed 
in Section \ref{subsec.dumbbell} contained two and three cells per individual. 
In this section, reticulated virtual creatures 
with a larger number of cells that make up one individual will be discussed.
Note that this reticulated virtual creature was observed in simulations 
performed independently of the simulation presented in Section \ref{subsec.dumbbell}.
The only difference between the two simulations is the seed value 
of the random number, which affected the formation of the terrain, 
mutations, and locations where cells and photons were generated.

\vspace{2mm}
\subsubsection{Number of connections and energy transport}
\label{sec.det.spec.01}

\begin{figure}[htbp]
    \centering
    \includegraphics[width=1.0\hsize]{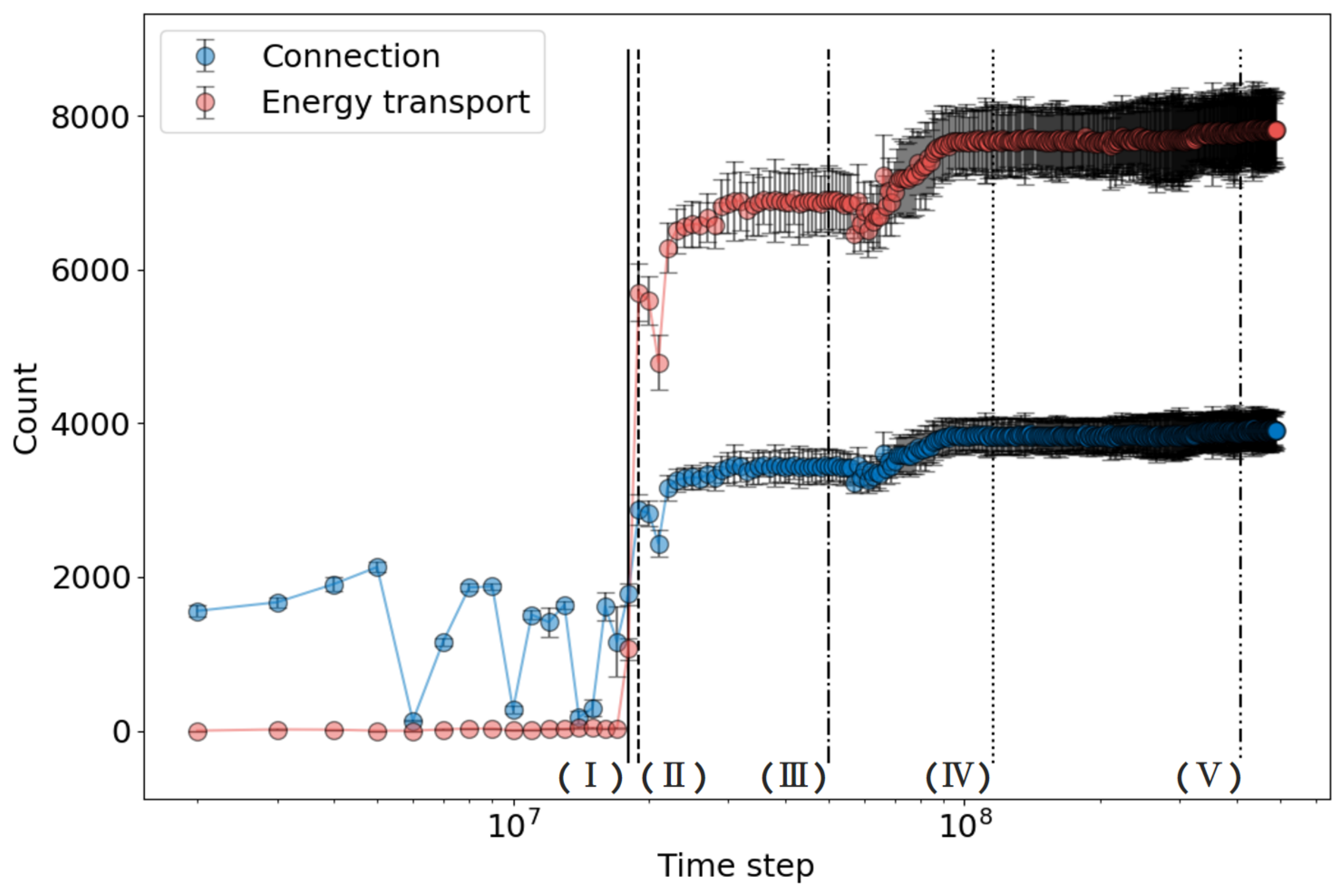}
    \caption{
    Variation of the number of connections and energy transport times 
    across the field versus simulation time. 
    The error bars indicate invariant standard deviations. 
    ($\mathrm{I}$), 
    ($\mathrm{I}\hspace{-1.2pt}\mathrm{I}$), 
    ($\mathrm{I}\hspace{-1.2pt}\mathrm{I} \hspace{-1.2pt}\mathrm{I}$), 
    ($\mathrm{I}\hspace{-1.2pt}\mathrm{V}$), 
    and ($\mathrm{V}$)
    are labels given to times with characteristic energy transport 
    and number of bonds. 
    }
    \label{fig.07}
\end{figure}

As noted in Section \ref{subsec.decay}, virtual creatures 
with a higher number of connections consume more energy.
In order to increase the number of cells that make up one individual, 
the virtual creature must acquire a mechanism for efficiently distributing 
energy throughout its body.
The reticulated virtual creature shown here has acquired such a mechanism.
The change in the number of connections and 
energy transports over the entire field versus simulation time is shown in Fig. \ref{fig.07}.

\vspace{2mm}

The energy transport between connected cells 
was approximately $0$ until near the simulation time 
($\mathrm{I}\hspace{-1.2pt}\mathrm{I}$), 
and the number of connections fluctuated widely between $0$ and $2000$. 

\begin{figure}[htbp]
    \centering
    \includegraphics[width=1.0\hsize]{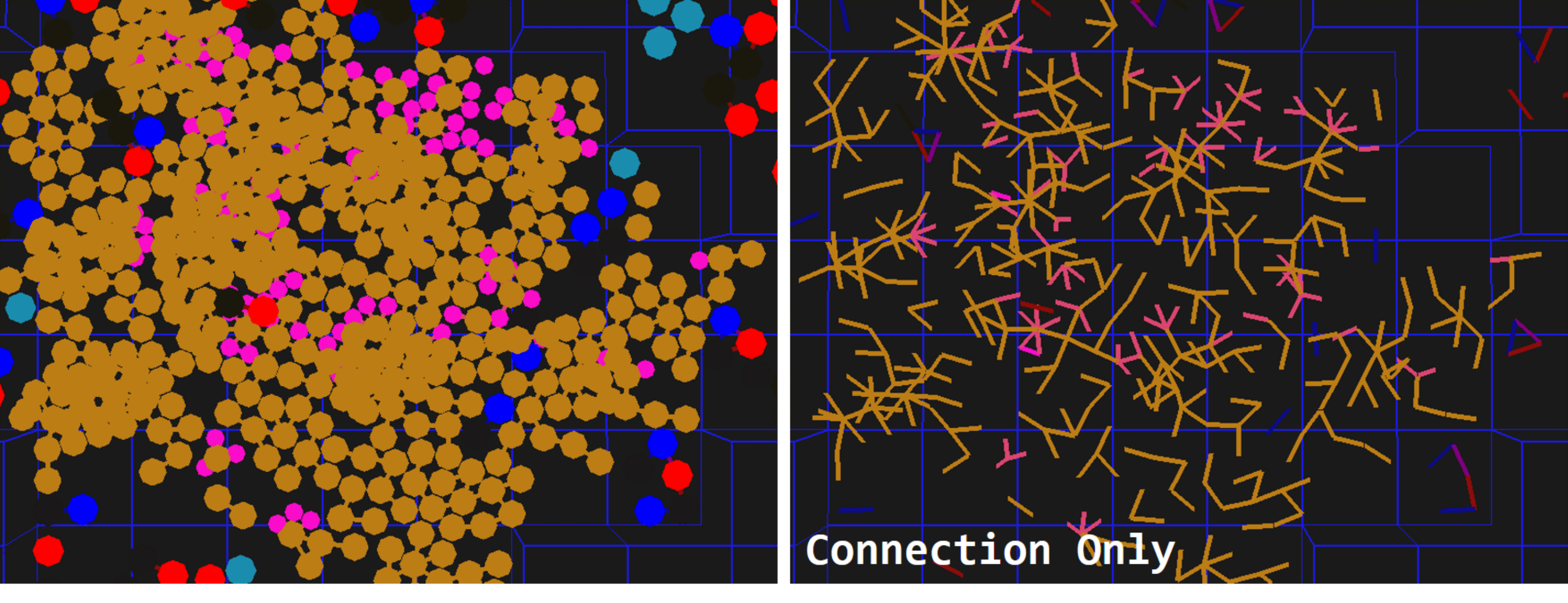}
    \caption{
    Example of a virtual creature from Fig. \ref{fig.07} ($\mathrm{I}$). 
    The left image shows a cell and its bonding, 
    while the right image shows only the bonding. 
    }
    \label{fig.08}
\end{figure}

This fluctuation may be because virtual creatures 
with many connections were born but could not stably survive 
because the energy was not distributed properly.
An example of a virtual creature from Fig. \ref{fig.07} ($\mathrm{I}$)
is shown in Fig. \ref{fig.08}. 
Only the bonding is visualized on the right side of the figure and
it has a more complex morphology than the dumbbell type.
It appears as if pink legs are growing against a yellow body, 
and these legs are being violently moved. 

\vspace{2mm}

\begin{figure}[htbp]
    \centering
    \includegraphics[width=1.0\hsize]{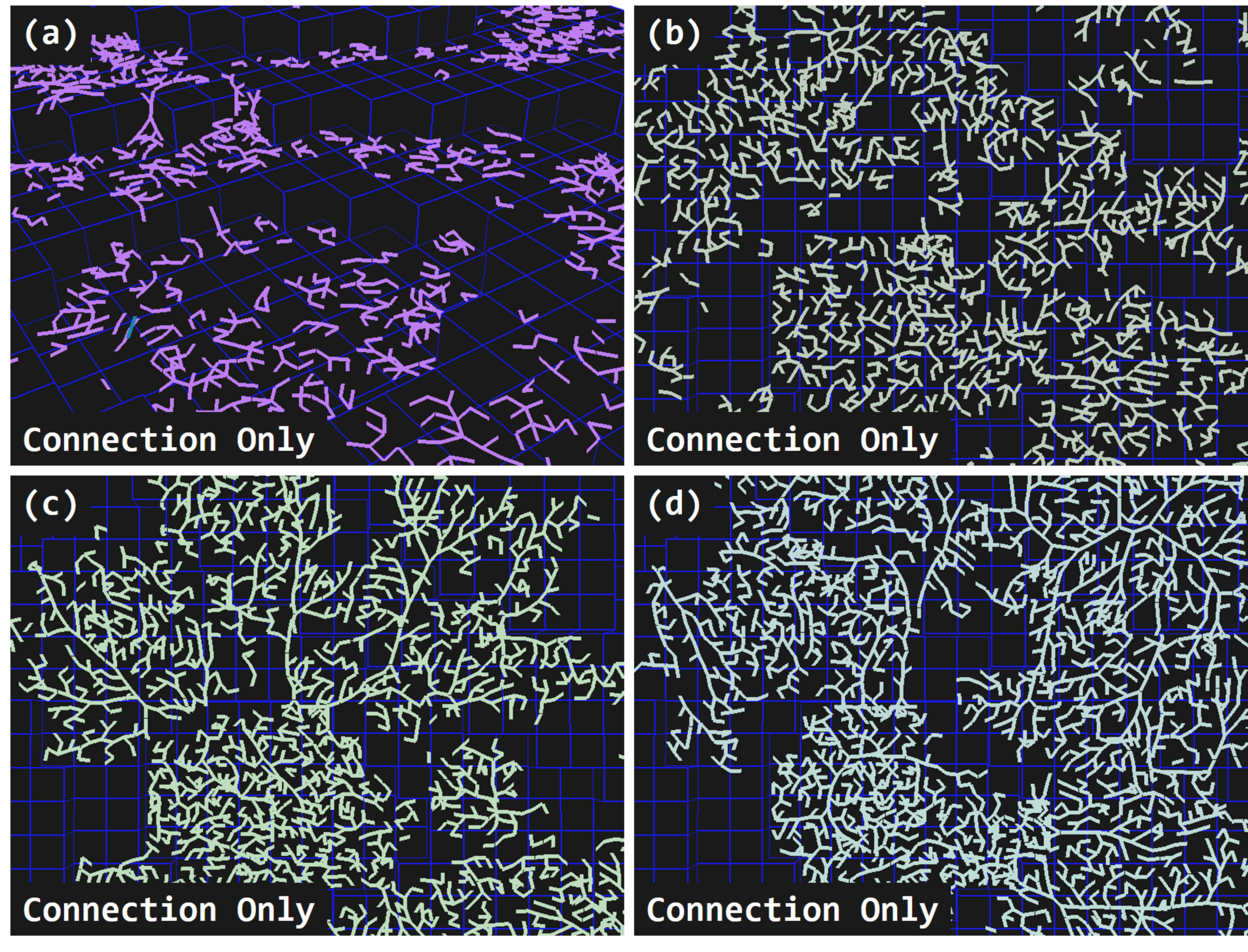}
    \caption{
    Virtual creatures seen on the field: (a), (b), (c), 
    and (d) represent times 
    ($\mathrm{I}\hspace{-1.2pt}\mathrm{I}$), 
    ($\mathrm{I}\hspace{-1.2pt}\mathrm{I} \hspace{-1.2pt}\mathrm{I}$), 
    ($\mathrm{I}\hspace{-1.2pt}\mathrm{V}$), 
    and ($\mathrm{V}$) from Fig. \ref{fig.07}, respectively. 
    The network grows larger and whiter in color 
    as the simulation time progresses. 
    }
    \label{fig.09}
\end{figure}

The number of energy transports increased in proportion 
to the number of connections at Fig. \ref{fig.07} ($\mathrm{I}\hspace{-1.2pt}\mathrm{I}$). 
This is the result of the virtual creatures 
obtaining the ability to distribute energy more efficiently. 
In fact, the number of connections was also stable from this point on.
The virtual creatures seen on the field at times 
($\mathrm{I}\hspace{-1.2pt}\mathrm{I}$), 
($\mathrm{I}\hspace{-1.2pt}\mathrm{I} \hspace{-1.2pt}\mathrm{I}$), 
($\mathrm{I}\hspace{-1.2pt}\mathrm{V}$), 
and ($\mathrm{V}$) from Fig. \ref{fig.07} are shown in Figs. \ref{fig.09} 
(a), (b), (c), and (d), respectively.
The network grew larger as simulation time progressed.
In addition, the color of the cells was accordingly closer to white, 
which had a small generation cost, as noted earlier.
This was considered as a survival strategy in which the energy that 
a single cell could acquire from light was reduced, thus creating 
a wide network and sharing of the energy across the entire network.

\vspace{2mm}
\subsubsection{How to spread offspring over a wide area}
\label{subsubsec.spread}

\begin{figure}[htbp]
    \centering
    \includegraphics[width=1.0\hsize]{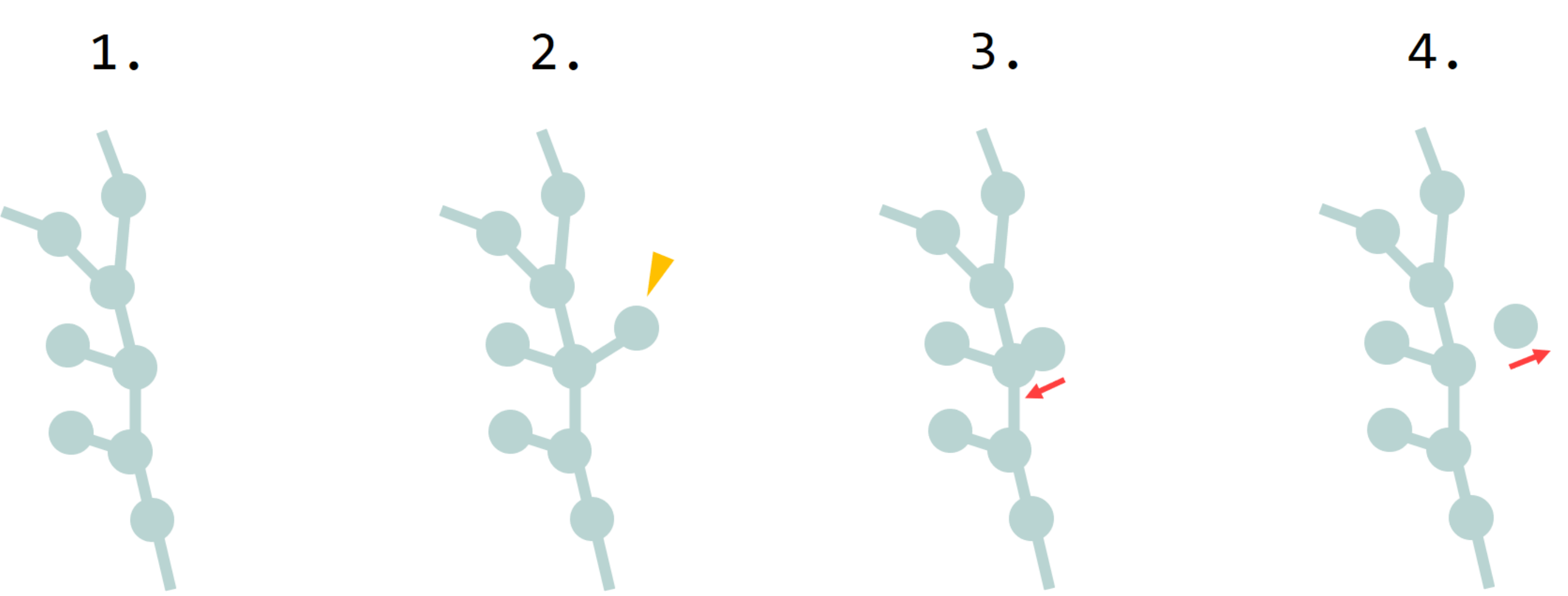}
    \caption{
    Schematic diagram of the cell disconnection: 
    Step 1 - original cell, 
    Step 2 - a fourth cell is expanded, 
    Step 3 - the connection distance between the expanded 
    and source cell is reduced, and Step 4 - the bond is broken
    when the reduction reaches the distance limit between the cells, 
    and the fourth cell is separated. 
    }
    \label{fig.16}
\end{figure}

This reticulated virtual creature had a maximum of three connections 
extending from a single cell.
Furthermore, it was expanding 
the fourth cell and disconnecting it simultaneously. 
A schematic diagram of this situation is shown in Fig. \ref{fig.16}.

\vspace{2mm}

The cells expanded in Step 2 are pulled back to the cell 
from which they were generated.
The natural length reduction of this spring is derived 
from the output signal of the neural network, 
which is described in Section \ref{subsec.nn}. 
Here, the natural length can be shortened to zero, 
but they cannot completely overlap each other because of the size of the cells. 
Therefore, contact between cells caused the spring to be stretched as the natural length of the spring was shortened.
This connection will eventually break 
because the bond is designed to break when the distance 
between cells is twice the natural length, as mentioned in Section \ref{sec.conn}.
Moreover, disconnected cells are separated by repulsion.
This mechanism is an essential function of this virtual creature 
to push offspring farther away.  

\vspace{2mm}

It is worth noting that this disconnection is not a behavior derived 
from the Book, it is caused by a neural network.
This is because this neural network-triggered disconnection 
can be achieved flexibly depending on the surrounding circumstances.
In this case, the disconnection was performed under the condition 
that the number of connections was four. 
The reticulate virtual creature could have acquired 
the survival strategy whereby reaching three bonds 
was enough for growth 
and thus it was better to move to the procreation phase.

\vspace{2mm}
\subsubsection{Energy transport and parameters $\Delta s$}
\label{sec.det.spec.01}

As mentioned earlier, the energy distribution was significant 
in a virtual creature composed of a large number of cells. 
The amount of energy to distribute was determined by the neural network 
presented in Section \ref{subsec.nn}.
How virtual creatures are affected 
when one parameter $\Delta s$ that determines the behavior 
of the neural network is changed will now be discussed.
The simulations here were run with $A = 2.0$, 
as discussed in Section \ref{subsec.decay}.

\vspace{2mm}

This reticulated virtual creature evolved and grew in 
an environment where $\Delta s = 0.1$.
First, the amount of available energy transport capacity 
in this setting was checked.

\vspace{2mm}

This experiment was performed using the following procedure. 

\vspace{2mm}

\begin{figure}[htbp]
    \centering
    \includegraphics[width=1.0\hsize]{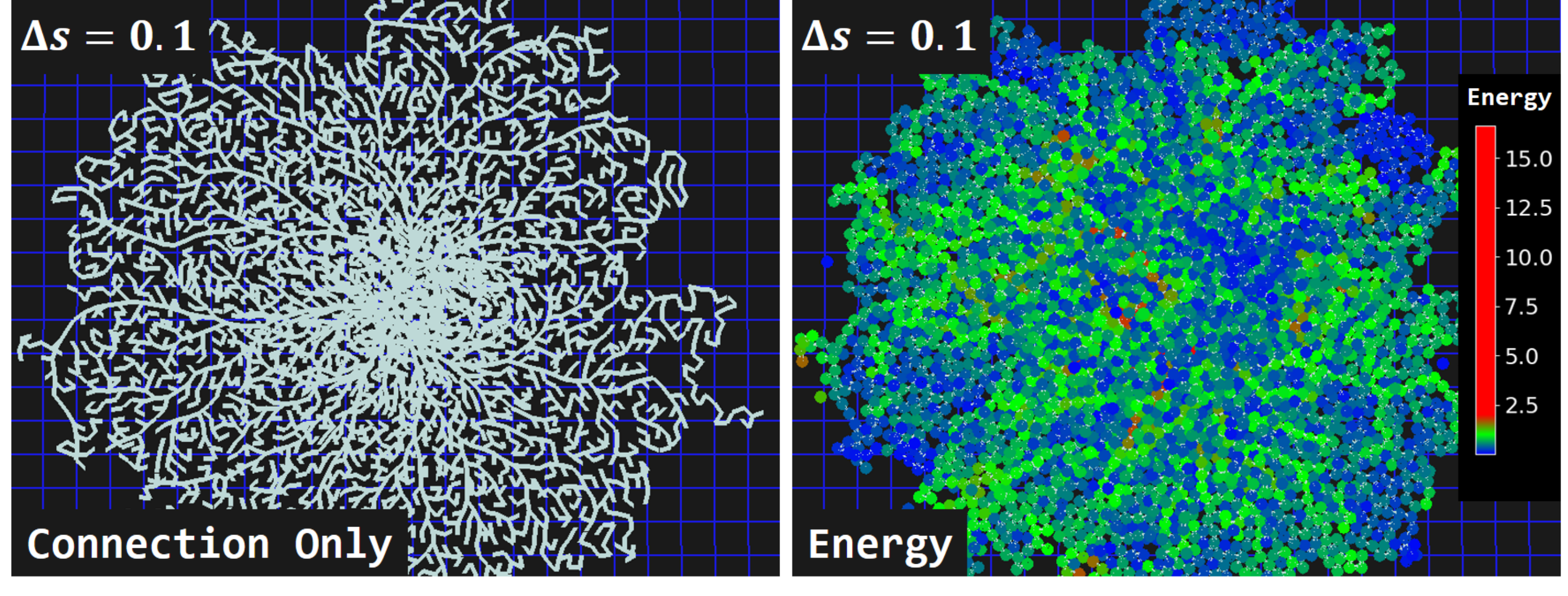}
    \caption{
    One cell from Fig. \ref{fig.09} ($d$) was sampled and fixed in the center of 
    the flat field, and the simulation was run with constant energy 
    applied only to that fixed cell. 
    The magnitude of this energy was set to twice the energy required 
    to produce the same cell. The left panel shows only the cell bonding, 
    while the right panel visualizes the energy of each cell. 
    $\Delta s$ was set to $0.1$.
    }
    \label{fig.17}
\end{figure}

First, one cell was sampled from Fig. \ref{fig.09} ($d$) and fixed 
in the center of the flat field.
Second, energy was continually provided only to that cell 
so that the cell continued to hold a certain amount of energy. 
This energy was set to be twice the energy 
required to produce the same cell.
Third, cutting off light from the sun was simulated, and only energy from this fixed cell was supplied to the other cells. 
Even if this fixed cell generates other cells, 
the generated cell will die 
if the fixed cell cannot successfully transport energy. 
The number of cells in the field equilibrated after the simulation was run for some time.
The reticulated virtual creatures at this time are 
shown in Fig. \ref{fig.17}. 

\vspace{2mm}

The reticulated virtual creature could form a large body composed of many cells, 
even though only a fixed central cell was supplied with energy.
This result indicates that energy transport was conducted efficiently.

\vspace{2mm}

\begin{figure}[htbp]
    \centering
    \includegraphics[width=1.0\hsize]{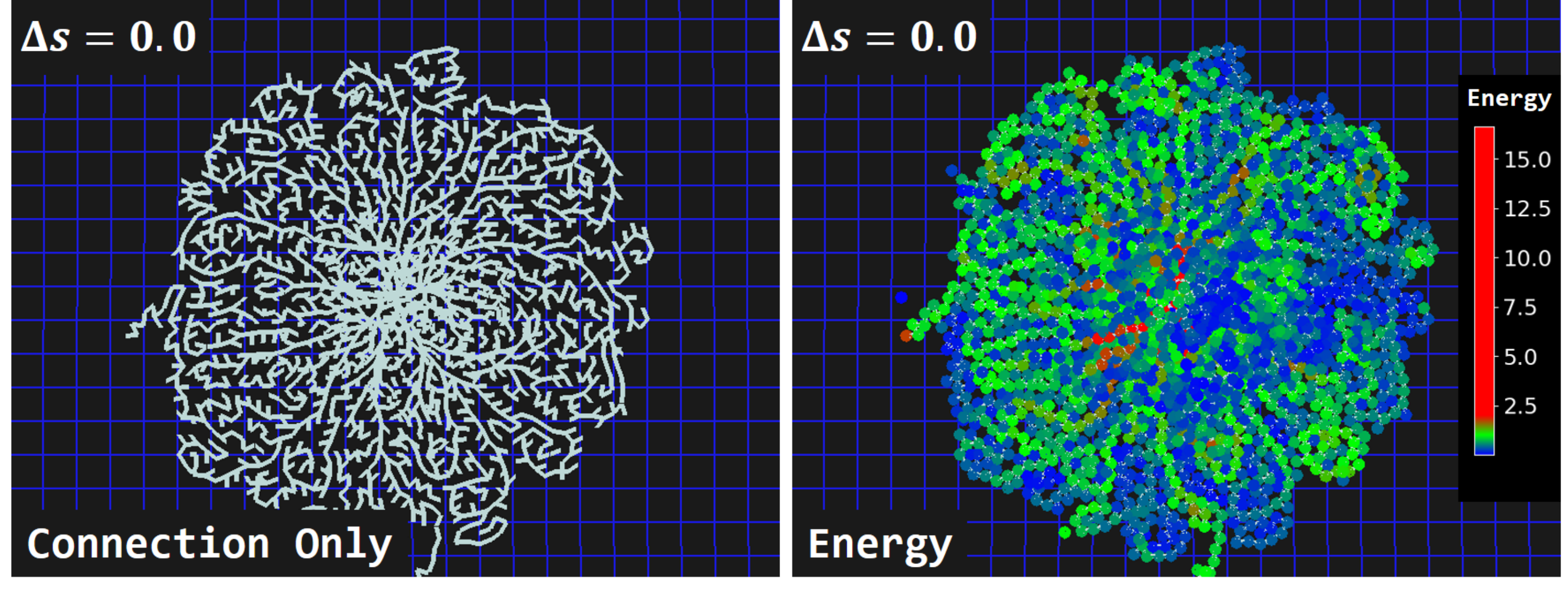}
    \caption{
    One cell from \ref{fig.09} ($d$) was sampled and fixed in the center of the field, 
    and the simulation was run with constant energy applied only to 
    that fixed cell. 
    The magnitude of this energy was set to twice the energy required to 
    produce the same cell. The left figure shows only the cell bonding, 
    while the right figure visualizes the energy of each cell. 
    $\Delta s$ was set to $0.0$.
    }
    \label{fig.18}
\end{figure}

Next, the simulation results when the parameter was changed 
to $\Delta s=0.0$ are shown in Fig. \ref{fig.18}.

\vspace{2mm}

Compared to Fig. \ref{fig.17}, the network spread was smaller 
and $\Delta s = 0.0$ caused problems in the
information transfer between cells, 
indicating that the energy transport did not work well.

\vspace{2mm}

\begin{figure}[htbp]
    \centering
    \includegraphics[width=1.0\hsize]{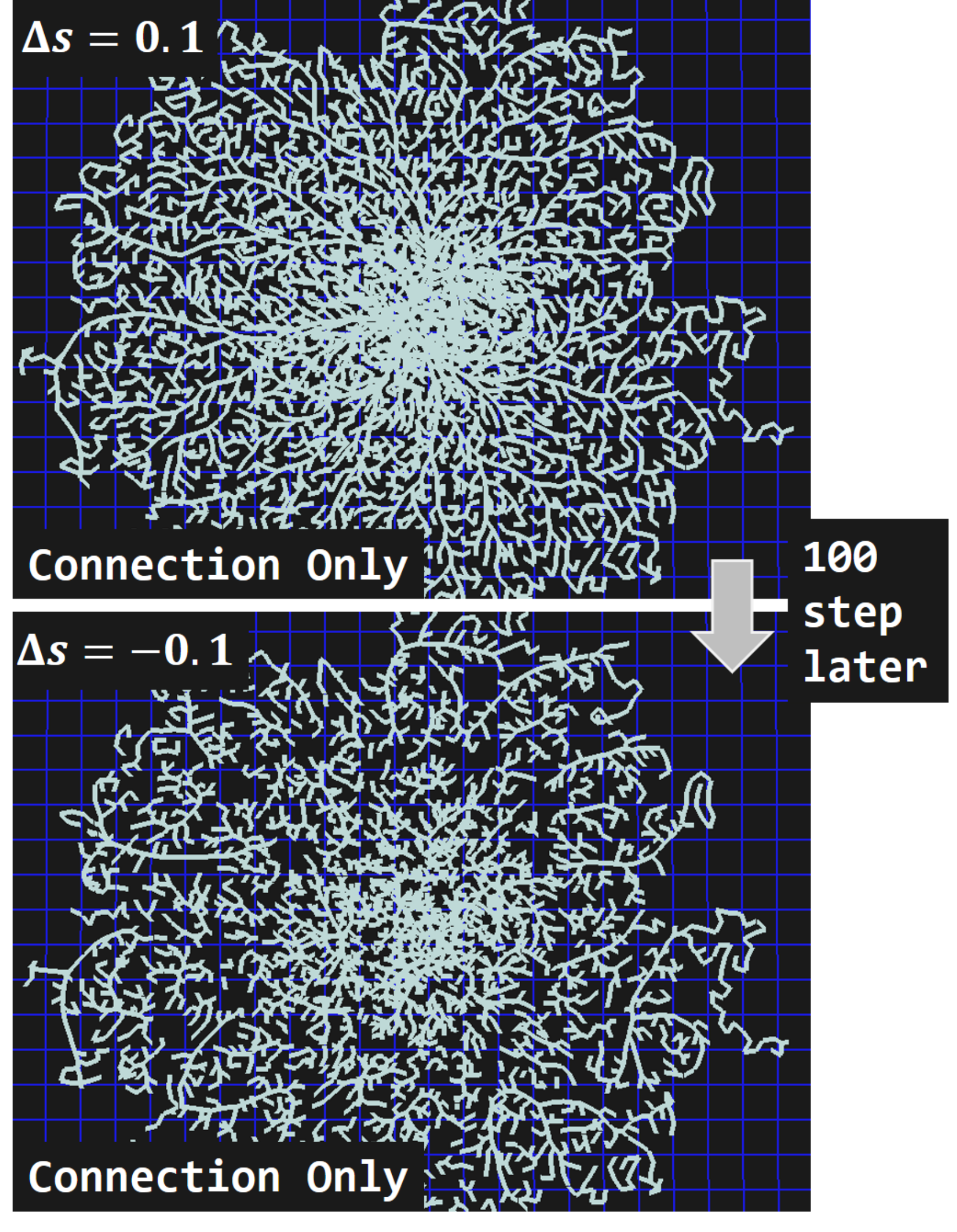}
    \caption{
    $\Delta s = -0.1$ for the reticulated virtual creature from Fig. \ref{fig.17}. 
    }
    \label{fig.12}
\end{figure}

Lastly, $\Delta s$ was set to $-0.1$ for the reticulated virtual creature from Fig.~\ref{fig.17}, and the
results are shown in Fig. \ref{fig.12}.

In this case, the reticulated virtual creatures were scattered 
because the change of $\Delta s=-0.1$ malfunctioned 
the diffusion process that was discussed in Section \ref{subsubsec.spread}. 

\vspace{2mm}

These results indicate that Hebb's rule, which was incorporated 
in a simplified manner, had a significant impact on energy transport 
and the diffusion of offspring.
Specifically, energy transport exhibited a problem and the size of the reticulated virtual creature was reduced if $\Delta s$ was set to $0.0$ and grown without updating 
the coupling strength.
In addition, providing inverse feedback on the coupling strength 
by setting $\Delta s$ to $-0.1$ caused a collapse 
that may be due to a malfunction of the offspring diffusion process. 
This result demonstrates the importance of cell-to-cell 
communication in reproduction and development.

\vspace{2mm}
\section{Conclusion}
\label{sec.conc}

This study proposed a new model of artificial life that could handle 
reproduction, development, and individual interactions in a composite way. Furthermore, two virtual creatures were introduced in this model. 

\vspace{2mm}

The reticulated virtual creature had an interesting way of reproduction 
in which it detached parts of its own body and separated far from them. 
The virtual creature also grew a body composed of many cells by efficiently sharing energy, 
which is a more advanced developmental process than that in 
earlier tetrahedral virtual creatures.
Regarding individual interactions, 
the dumbbell-shaped virtual creature acquired a form of protection 
by expanding a protective organ during division, 
making itself less susceptible to predation by other virtual creatures. 

\vspace{2mm}

However, several properties were not seen in this simulation. 
For example, all the virtual creatures seen in this study were 
self-reproductions through asexual reproduction.
Additionally, there was no evolution of vertical growth or active movement.
It was assumed that one of the causes for this was simply the short simulation time, 
but the possibility that the model was not designed 
to easily produce such virtual creatures cannot be ruled out. 
Another problem was that the field used in this simulation was 
not large enough compared to the size of the cell.
Therefore, there were almost exclusively the same virtual creatures in the same field as the simulation proceeded. 
In addition, the effects of field undulation geometry on evolution 
and changes in Book length over the course of evolution were not researched.
Further studies are needed on these points.

\section{Acknowledgments}

I would like to thank Hirokazu Nitta for his feedback, 
as it greatly helped improve the paper as a whole, 
especially the conclusion section.
The advice and comments provided by Soichi Ezoe have also been a great help in 
improving the methodology section. 
I am also deeply grateful to Kazumasa Itahashi. 
Kentaro Yonemura also read the paper carefully and made some helpful comments, 
especially about the wording on the paper. 

\section* {Appendix}
\label{sec.app}

The energy $E$ of the cell after $\Delta t$ seconds 
(one step) is updated as follows: 

\begin{align}
E\left(t+\Delta t\right) 
&= 
E\left(t\right) - 
\left(
C \cfrac{A^{N_c}}{A^{N_{\rm max}} }
\right)
E\left(t\right) {\Delta t}\nonumber\\
&=
E\left(t\right) - 
U
E\left(t\right)
{\Delta t} \quad ,
\end{align}

where $N_{\rm max}$ is the maximum number of cells that 
can be connected to the cell 
and $N_c$ is the number of cells that are currently connected to the cell. 

\vspace{2mm}

This can also be expressed as:

\begin{align}
\cfrac{E\left(t+\Delta t\right) - E\left(t\right)}{\Delta t}
&=  - U E\left(t\right) .
\end{align}

If $\Delta t \to 0$, 
the above equation can be written as: 

\begin{align}
\cfrac{d}{d t} E\left(t\right)
&=  - U E\left(t\right) . 
\end{align}

This equation can be easily solved, 
which results in:

\begin{align}
E\left(t\right)
&=  E_0 e^{-Ut} ,
\end{align}

where $E_0$ denotes $E\left(0\right)$.

\vspace{2mm}

The energy obtained from a photon is calculated 
from the intensity $I$ of the light, light absorption rate $a$ 
of the cell receiving the light, 
and energy conversion efficiency $c$ of the light, which is expressed as:

\begin{align}
\Delta E_L = c_r I_r a_r + c_g I_g a_g + c_b I_b a_b \quad, 
\end{align}

where the subscripts $r$, $g$, and $b$ refer to 
the red, green, and blue components, respectively.
Therefore, when a single cell receives a photon at intervals of $T$ 
seconds on average, the change in the energy of that cell is expressed as:

\begin{align}
E\left(\left(n+1\right)T\right) 
= 
E\left(nT\right) e^{-UT}
+\Delta E_L .
\label{ap.eq.ns}
\end{align}


\vspace{2mm}

Here, the numerical sequence is considered, which is expressed as:

\begin{align}
f_{n+1} = r f_n + b \quad . 
\end{align}

The following equation is obtained as a general term: 

\begin{align}
f_n = \left(f_0-\cfrac{b}{1-r}\right)r^n + \cfrac{b}{1-r} 
\quad ,
\label{ap.eq.ns2}
\end{align}

where $|r|<1$. 

If $E\left(t\right)$ is regarded as a sequence of numbers defined by 
the graded Eq. (\ref{ap.eq.ns}), 
and $f_n=E\left(nT\right), r=e^{-UT}, b=\Delta E_L$ 
in Eq. (\ref{ap.eq.ns2}), then: 

\begin{align}
E\left(nT\right) = 
\left(E_0-\cfrac{\Delta E_L}{1-e^{-UT}}\right)e^{-nUT} + 
\cfrac{\Delta E_L}{1-e^{-UT}} \quad ,
\end{align}

where $0<UT$. Therefore, $e^{-UT}<1$. 

\vspace{2mm}

$E_{\infty}$ can be estimated as energy when enough time 
has passed and by using $n\to\infty$, 
which is expressed as:

\begin{align}
E_{\infty} =  
\cfrac{\Delta E_L}{1-e^{-UT}} \quad .
\end{align}


\vspace{2mm}

Next, $T$ is calculated.

\vspace{2mm}

The emitting cell releases an amount of light proportional 
to the square of its radius $R$ during a unit time. 
Thus, the amount of light $P$ received per unit area 
during a unit time at a distance $h$ is expressed as:

\begin{align}
P=\cfrac{DR^2}{4\pi h^2} .
\end{align}


\vspace{2mm}

If the area over which a cell receives light is $\Delta S$, 
then the amount of light that a cell receives per unit time 
can be expressed as $P\Delta S$.
This value is smaller than $1$ in the simulations performed here. Therefore, it can be regarded as the probability $p$ of a cell 
receiving one light per unit time, which is expressed as:

\begin{align}
p = P\Delta S .
\end{align}

\vspace{2mm}

Subsequently, the average time $T$ that a photon hits can be expressed as: 

\begin{align}
T 
&= \sum_{n=1}^{\infty} n\left(1-p\right)^{n-1} p = \cfrac{1}{p} .
\end{align}


\vspace{2mm}

Finally, the following is obtained:

\begin{align}
E_{\infty} =  
\cfrac{\Delta E_L}{1-\exp\left[{-\cfrac{U}{P\Delta S}}\right]} .
\end{align}


\bibliographystyle{apsrev4-1}
\bibliography{references.bib}
\end{document}